\documentclass[journal]{IEEEtran}

\usepackage{cite}
\usepackage{amsmath,amssymb,amsfonts}
\usepackage{graphicx}
\usepackage{placeins}
\usepackage{booktabs}
\usepackage{multirow}
\usepackage{rotating}
\usepackage{url}
\def\BibTeX{{\rm B\kern-.05em{\sc i\kern-.025em b}\kern-.08em
    T\kern-.1667em\lower.7ex\hbox{E}\kern-.125emX}}
\usepackage[colorlinks=true,linkcolor=blue,citecolor=blue,urlcolor=blue,breaklinks=true]{hyperref}
\newcommand{\dcite}[1]{\textsuperscript{\cite{#1}}}

\begin{document}

\renewcommand{\topfraction}{0.95}
\renewcommand{\bottomfraction}{0.9}
\renewcommand{\textfraction}{0.06}
\renewcommand{\floatpagefraction}{0.85}
\renewcommand{\dbltopfraction}{0.95}
\renewcommand{\dblfloatpagefraction}{0.85}
\setlength{\textfloatsep}{6pt plus 2pt minus 2pt}
\setlength{\floatsep}{6pt plus 2pt minus 2pt}
\setlength{\intextsep}{6pt plus 2pt minus 2pt}
\setlength{\dblfloatsep}{6pt plus 2pt minus 2pt}
\setlength{\dbltextfloatsep}{6pt plus 2pt minus 2pt}

\title{Coding What Matters: A Semantic-Aware Memory Interface for
Energy-Efficient Perception in Autonomous Vehicles}

\author{Haohua~Que and Handong~Yao\textsuperscript{*}%
\thanks{Manuscript received Month Day, Year; revised Month Day, Year.
\textsuperscript{*}Corresponding author. \textit{(Corresponding author: Handong Yao.)}}
\thanks{The authors are with the College of Engineering, University of Georgia,
Athens, GA 30602 USA (e-mail: Haohua.Que@uga.edu; Handong.Yao@uga.edu).}}

\markboth{IEEE Transactions on Intelligent Transportation Systems,~Vol.~XX, No.~X, Month~Year}%
{Que and Yao: MotiMem-$\Omega$: Semantic-Aware Memory-Interface Coding}

\maketitle

\begin{abstract}
Autonomous vehicles stream high-resolution surround-camera frames into memory before perception runs. This
sensor-to-memory path consumes energy when cells store ones and adjacent bytes toggle on the data bus, so its
cost follows bit-1 density and switching activity rather than pixel semantics. We present
\mbox{MotiMem-$\Omega$}, a semantic-aware memory-interface coder that lowers this cost while preserving
perception predictions. Its semantic importance field protects traffic participants, especially vulnerable road
users, while assigning lower fidelity to sky and empty background. Cross-dataset bit-sensitivity sweeps determine
class weights, with a safety floor for pedestrians, cyclists, and motorcyclists. Each image block then selects a
precision tier by minimizing a joint energy--distortion cost. When ego pose is available, a motion-compensated
prior carries protected regions between frames. We estimate interface-energy reduction from the two measured
proxies using a coefficient-swept memory-energy model. Across 29 detectors on 12 driving datasets, 5 occupancy
models, and 5 segmentation networks, \mbox{MotiMem-$\Omega$} retains about 90\% of detection mean average
precision, 91\% of vulnerable-road-user recall, over 98\% of occupancy accuracy, and the strongest segmentation
retention among energy-reducing methods. It reduces front-camera bit-1 density by 52\%, corresponding to a
modeled memory-interface energy reduction near 36\%, with a lower end of 27\% under the literature coefficient
sweep. It also gives higher retention than the baseline and energy-matched truncation at the same or lower bit-1
density, whereas image codecs preserve accuracy without reducing memory-interface energy.
\end{abstract}


\IEEEpeerreviewmaketitle

\section{Introduction}
\begin{figure}[!t]
\centering
\includegraphics[width=\columnwidth]{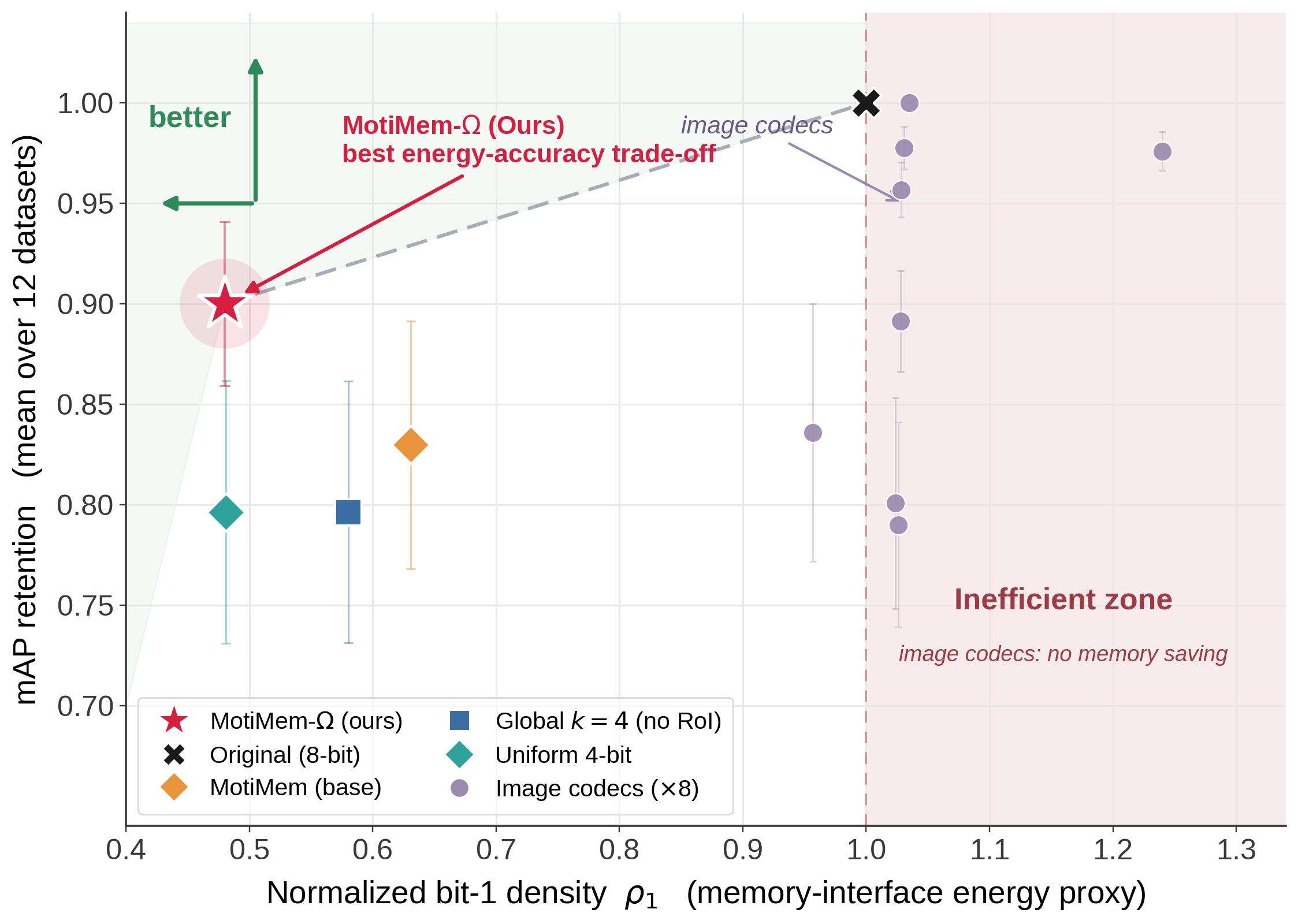}
\caption{Detection accuracy versus memory-interface energy across the 12 driving datasets. The vertical axis is
mAP retention (each of the 29 detectors measured against its own original-frame detections, then averaged over
the 12 datasets); the horizontal axis is normalized bit-1 density $\rho_1$, our memory-interface energy proxy,
on the same 12-dataset front-camera basis as Table~\ref{tab:det}. MotiMem-$\Omega$ gives the strongest
accuracy--energy trade-off among the energy-reducing methods, retaining about $90\%$ of mAP at
$\rho_1\!\approx\!0.48$, where
energy-matched truncation (uniform 4-bit, global $k{=}4$) and the baseline give up $7$ to $10$ points
of accuracy at the same or higher energy, while image codecs lie at $\rho_1\!\ge\!1$ with no memory saving.}
\label{fig:teaser}
\end{figure}
\IEEEPARstart{M}{odern} autonomous vehicles carry several high-resolution cameras that generate raw frames at
tens of frames per second. Before any neural network runs, each frame is written to a DRAM or SRAM frame buffer
and read back again, and this traffic continues for as long as the vehicle operates. The energy of that
sensor-to-memory path is spent charging memory cells and toggling the wires of the data bus, and it grows with
two properties of the stored bytes: how many bits are set to one, and how often adjacent bits change
\cite{stanburleson,jedec}. These two byte-level quantities are not the same as the semantic value of a pixel. A
pedestrian, a car, the road surface, and the sky can have very different importance for driving, even when their
stored bytes cost the memory interface through the same physical mechanisms. As the number of cameras and their resolution keep
increasing, this interface energy becomes a real share of the sensing power \cite{linasplos,fengtits}, yet most
efficiency work optimizes network compute or compressed file size rather than the interface energy.

Existing methods leave a gap between low-energy storage and scene-aware perception. Efficient perception methods,
including quantization, pruning, and compact backbones \cite{quant,pruningli,mobilenetv2,efficientnet}, reduce
network compute and parameter movement after the frame has already been stored. They do not reduce the
sensor-to-memory traffic of the sensing path. Coding-for-machines and region-of-interest compression
\cite{vcm,vcmyang,jpegai,roinvc} preserve task accuracy while shrinking files, but file size is not a memory
interface metric: entropy-coded bytes do not explicitly minimize bit-1 density or switching and can raise bus
toggling. Approximate-memory and selective bit-dropping schemes \cite{jetcas,raha,approxsram,eden} do target
memory energy, but they usually apply the same coarseness to all content. They do not know whether a byte belongs
to a pedestrian, a vehicle, the road, or the sky. The missing piece is therefore a memory-interface coder that is
both energy-aware and scene-aware. It should reduce bit-1 density and switching, preserve perception outputs,
protect safety-critical road users, and hold across the detectors, dense-perception models, and datasets that an
autonomous vehicle will meet.

In this paper we propose \mbox{MotiMem-$\Omega$}, which improves the accuracy--energy trade-off by giving the
memory interface a coarse understanding of the traffic scene. A human driver does not inspect every pixel with
the same priority, and neither should an approximate frame buffer. Traffic participants carry decision-critical
evidence; vulnerable road users require the strongest protection; sky and empty background are usually safe
places to save energy; and road or ground regions are coded at a conservative, higher precision because distant
or not-yet-detected road users can appear on them. MotiMem-$\Omega$ encodes this intuition as a semantic
importance field.

The field converts detections and a sky prior into block-level memory importance. Its high-priority classes
include pedestrians, cyclists, motorcyclists, and vehicles; its lower-priority regions include empty background
and sky; and road/ground context is not aggressively truncated. Class weights are learned from cross-dataset
bit-sensitivity sweeps, with a safety floor for vulnerable road users. Each image block then chooses its own
precision by solving a local energy--distortion problem, so most of the saving comes from low-importance regions
while object regions remain near-lossless. A motion-compensated prior carries protected regions from one frame to
the next when ego pose is available. The earlier conference version (baseline), which coded one region of
interest at one precision, is the two-level special case of the present semantic and hierarchical
scheme~\cite{motimem}. The contributions of this paper are as follows.
\begin{itemize}
\item We propose a semantic importance field that converts detections and a sky prior into block-level memory
importance, together with a safety-aware hierarchical coder that assigns each region an energy-optimal
precision. Together, they create a scene-aware memory interface that reduces bit-1 density and switching
activity along the camera-to-memory path, lowering interface energy while preserving high perception accuracy.
\item We present a comprehensive evaluation that tests 29 detectors
\cite{yolo,rtdetr,rtdetrv2,dfine,deim,dino,codetr} on 12
geographically diverse driving datasets
\cite{nuscenes,waymo,av2,once,pandaset,cadc,bdd100k,a2d2,carina,usyd,kuas,jodd}, together with 5
state-of-the-art occupancy models \cite{opus,sparseocc,daocc,gaussrender,gaussianformer2} and 5 segmentation
networks \cite{deeplabv3,fcn,lraspp,segformer,mask2former}, against about ten encoding baselines
(Fig.~\ref{fig:teaser}).
\end{itemize}
\begin{figure*}[t]
\centering
\includegraphics[width=\textwidth]{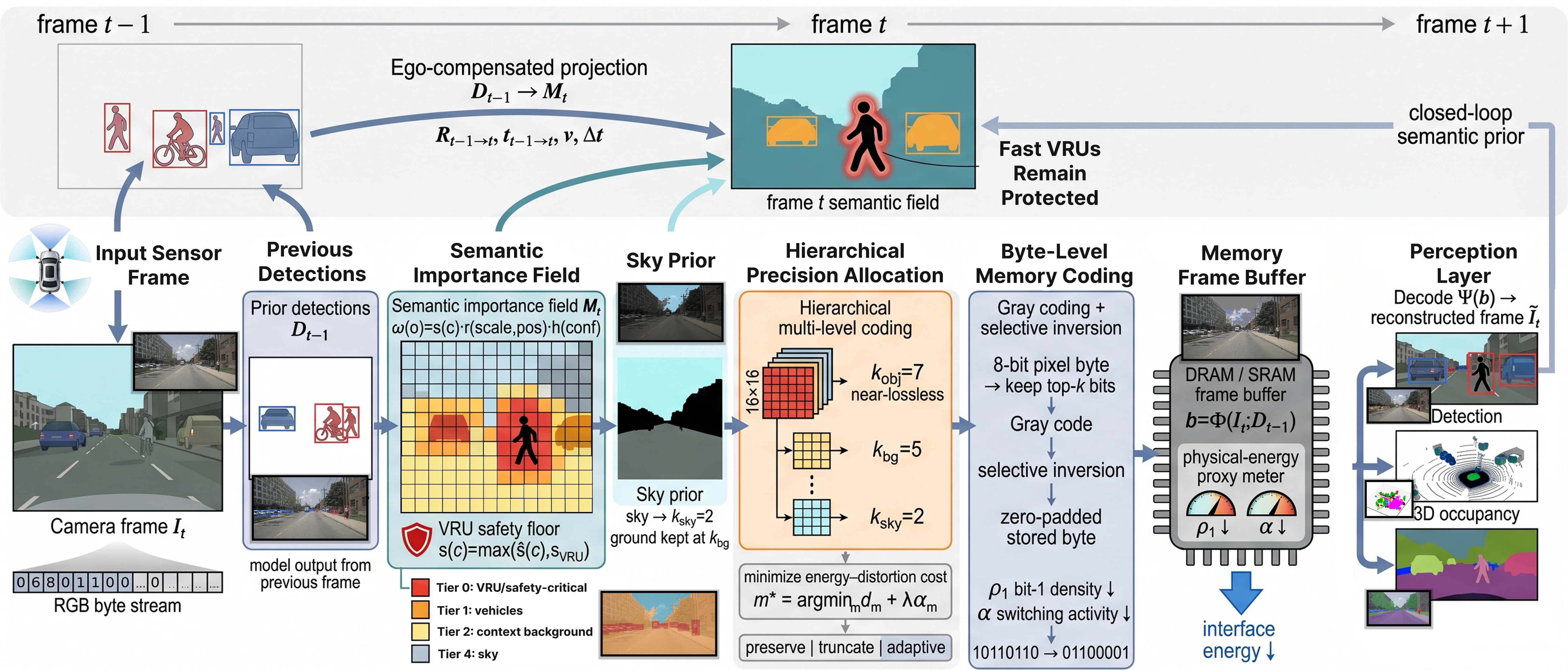}
\caption{Overview of the MotiMem-$\Omega$ encoding pipeline: $t{-}1$ detections $\rightarrow$ semantic importance
field $\rightarrow$ tiered hierarchical coding ($+$ sky prior) $\rightarrow$ DRAM/SRAM frame buffer $\rightarrow$
decode $\rightarrow$ perception; the field is projected $t{-}1\!\rightarrow\!t$ by ego-compensated motion.}
\label{fig:overview}
\end{figure*}

\section{Related Work}

\subsection{Approximate and Low-Power Memory}
The bit-1 density and switching objectives used by MotiMem-$\Omega$ are grounded in low-power memory and
interconnect research. Bus-invert coding \cite{stanburleson} and Gray-coded addressing
\cite{graysu} cut the number of bit transitions on a bus, and the same principle is deployed in industry as
data-bus inversion in DDR4 \cite{jedec}. Approximate computing relaxes exactness for energy, and Mittal's survey
\cite{mittalapprox} outlines where such relaxation is safe. At the storage level, approximate memory trades
reliability for power: quality-configurable DRAM scales the refresh rate per region \cite{raha}, and
significance-aware SRAM protects only the most important bits \cite{approxsram}. Voltage scaling characterized on
real chips \cite{voltron} and approximate DRAM applied to network inference \cite{eden} push memory energy lower
still. The primitive closest to ours is selective bit dropping with most-significant-bit
inversion and a least-significant-bit flag \cite{jetcas}, which our hierarchical coder generalizes. A parallel
literature shows that data movement, not arithmetic, dominates deep-network energy: accelerators such as Eyeriss
\cite{eyeriss}, EIE \cite{eie}, and the TPU \cite{tpu}, the Deep Compression pipeline \cite{deepcompression},
and the tutorial of Sze et al. \cite{szesurvey} all reduce or reuse data to save power. These methods either
treat every byte uniformly or operate inside the network; none reshapes the camera bitstream by where the
important content lies before it reaches the frame buffer.

\subsection{Compression and Coding for Machine Vision}
A second line compresses images and video for downstream machines rather than for human viewing. The Video
Coding for Machines effort \cite{vcm,vcmyang} and the JPEG~AI standard \cite{jpegai} target both human
and machine consumption, and learned codecs for machines \cite{icmle,taskcomp} optimize the bitstream against a
recognition loss. The broader neural-compression literature, from the end-to-end transforms of Ball\'e et al.
\cite{balle18} to autoregressive and Gaussian-mixture entropy models \cite{minnen18}, recurrent
codecs \cite{toderici17}, and learned video compression \cite{dvc}, has matched or surpassed classical
codecs in rate-distortion. Region-of-interest and scalable approaches allocate more bits to salient regions or
to a machine base layer \cite{roinvc,scalablechoi}. These methods preserve accuracy and
shrink the file, but a smaller file is not lower memory-interface energy: an entropy-coded stream keeps bit
density near one half and raises switching, and studies of image quality on networks \cite{dodge16} confirm
that aggressive coding eventually harms accuracy. The gap is therefore not another file codec, but a storage-path
representation whose objective is the byte pattern written to memory: lower bit density and lower switching
without an entropy coder on the frame-buffer path.

\subsection{Efficient and Robust Perception for Autonomous Driving}
A third line makes the perception model itself cheaper. Quantization \cite{quant}, pruning and
distillation \cite{pruningli,distill}, and efficient backbones
\cite{mobilenet,mobilenetv2,efficientnet} reduce compute and parameter memory, while
energy-efficient accelerators \cite{eyeriss,szesurvey} and analyses of the autonomous-driving compute budget
\cite{linasplos,katareedge,murshededge} target the inference cost. A separate strand studies the robustness of
perception to image corruptions and adverse weather \cite{hendrycksc,michaeliswinter}, which is relevant because
any coding scheme perturbs the input. Surveys of autonomous-driving perception \cite{fengtits,grigorescu} and of
vulnerable-road-user detection \cite{caopedestrian} frame the accuracy and safety requirements. This work
optimizes or stress-tests the network; it leaves the sensor-to-memory data path, and the safety-aware
allocation of fidelity within a frame, unaddressed.

Taken together, these lines leave three research gaps. The target should be the camera-to-memory interface,
not only network compute or compressed-file rate. The allocation should be scene-aware, so safety-critical road
users and low-value background are not treated alike. And the validation should cover heterogeneous detectors,
dense-perception tasks, and datasets rather than one model family. MotiMem-$\Omega$ is designed for this gap: it
reshapes the stored camera bytes to lower bit-1 density and switching, while using a data-driven semantic field
and a VRU safety floor to decide where precision is spent.
  
\section{Problem Formulation}
Table~\ref{tab:notation} lists the symbols used throughout. At time $t$ a camera produces a frame
$\mathbf{I}_t\in\mathbb{R}^{H\times W\times 3}$. Before any perception model reads it, the frame is written to a
DRAM/SRAM frame buffer as a byte stream, held, and read back. We insert an encoder $\Phi$ that reshapes this
stored stream, $\mathbf{b}=\Phi(\mathbf{I}_t;\mathcal{D}_{t-1})$, conditioned on the detections
$\mathcal{D}_{t-1}$ produced on the previous frame, and a decoder $\Psi$ reconstructs
$\tilde{\mathbf{I}}_t=\Psi(\mathbf{b})$ on read-out. The energy of moving and holding $\mathbf{b}$ at the
interface is captured by two data-dependent quantities, the bit-1 density and the switching activity,
\begin{equation}
\begin{aligned}
\rho_1(\mathbf{b})&=\frac{1}{8N}\sum_{i=1}^{N}\operatorname{popc}(b_i),\\
\alpha(\mathbf{b})&=\frac{1}{8N}\sum_{i=2}^{N}\operatorname{popc}(b_i\oplus b_{i-1}),
\end{aligned}
\end{equation}
where $N$ is the number of stored bytes, $b_i$ the $i$-th byte, $\operatorname{popc}(\cdot)$ counts set bits,
and $\oplus$ is bitwise XOR; $\rho_1$ drives cell-array and data-bus-inversion energy while $\alpha$ drives
I/O-bus toggling energy. We write the interface-energy proxy as
\begin{equation}
E(\mathbf{b})=w_1\,\frac{\rho_1(\mathbf{b})}{\rho_1^{0}}+w_\alpha\,\frac{\alpha(\mathbf{b})}{\alpha^{0}}+w_0,
\label{eq:energy}
\end{equation}
where $\rho_1^{0}$ and $\alpha^{0}$ are the values of the uncoded frame and $(w_1,w_\alpha,w_0)$ are
energy-decomposition coefficients whose plausible ranges are taken from the memory-energy literature
(Sec.~\ref{sec:energy}). Let $g$ be any perception model and let
$\rho\!\left(g(\tilde{\mathbf{I}}_t),g(\mathbf{I}_t)\right)\in[0,1]$ denote the \emph{retention}, the agreement
between predictions on the decoded frame and on the original frame. MotiMem-$\Omega$ solves
\begin{equation}
\min_{\Phi}\; E\!\left(\Phi(\mathbf{I}_t;\mathcal{D}_{t-1})\right)\quad
\text{s.t.}\quad \rho\!\left(g(\tilde{\mathbf{I}}_t),g(\mathbf{I}_t)\right)\ge 1-\varepsilon\;\;\forall g,
\label{eq:problem}
\end{equation}
where $\varepsilon$ is the tolerated accuracy loss. Since the constraint must hold for \emph{every} model and
class, $\Phi$ cannot be tuned to one detector; it must instead allocate fidelity by a model-agnostic,
safety-aware notion of which pixels matter.

\begin{table}[!ht]
\centering
\caption{Primary notation.}
\label{tab:notation}
\small
\begin{tabular}{@{}ll@{}}
\toprule
Symbol & Meaning \\
\midrule
$\mathbf{I}_t,\tilde{\mathbf{I}}_t$ & original / decoded frame at time $t$ \\
$\Phi,\Psi$ & MotiMem-$\Omega$ encoder / decoder \\
$\mathcal{D}_{t-1}$ & detections on the previous frame \\
$\rho_1,\alpha$ & bit-1 density / switching activity \\
$E(\mathbf{b})$ & memory-interface energy proxy, Eq.~\eqref{eq:energy} \\
$\rho(\cdot),\varepsilon$ & retention / tolerated accuracy loss \\
$\omega(o)$ & importance of detection $o$ \\
$s(c),r,h$ & class / scale--position / confidence terms \\
$\mathbf{M}_t$ & $B\times B$ block importance field \\
$\ell,\,k_\ell$ & tier index / its bit precision \\
$m,\lambda$ & block coding mode / distortion--switching trade-off \\
$a_\ell,w_\ell$ & tier area fraction / importance weight \\
$D(k),D_{\max}$ & distortion at precision $k$ / distortion budget \\
$w_1,w_\alpha,w_0$ & energy-model coefficients \\
\bottomrule
\end{tabular}
\end{table}

\section{The \texorpdfstring{MotiMem-$\Omega$}{MotiMem-Omega} Method}
Fig.~\ref{fig:overview} gives an overview of MotiMem-$\Omega$. The pipeline has three stages: semantic-prior
construction (input frame, previous detections, motion projection, and semantic field), scene-prior precision
allocation (sky prior and hierarchical tiers), and the byte-level memory interface (stored coding, frame-buffer
energy meter, decode, and closed-loop perception).

\subsection{Semantic Prior Construction}
\noindent\textbf{Input Frame and Previous Detections.}
At time $t$, the encoder receives the camera frame $\mathbf{I}_t$ as an RGB byte stream and the detector outputs
$\mathcal{D}_{t-1}$ from the previous decoded frame. This is the left side of Fig.~\ref{fig:overview}: the current
sensor frame provides the bytes to be stored, while $\mathcal{D}_{t-1}$ provides a scene prior for where precision
should be protected.

\noindent\textbf{Ego-Compensated Projection.}
\label{sec:loop}
The pipeline starts with the ``Previous Detections'' and top closed-loop arrow in Fig.~\ref{fig:overview}. The
field at $t$ is built from detections at $t-1$, which are stale because the ego vehicle and the objects both
move. For each prior detection we back-project its ground-contact point to a 3D point $\mathbf{p}_{t-1}$,
extrapolate it with an ego-compensated constant-velocity model,
\begin{equation}
\mathbf{p}_t=\mathbf{R}_{t-1\to t}\big(\mathbf{p}_{t-1}+\mathbf{v}\,\Delta t\big)+\mathbf{t}_{t-1\to t},
\label{eq:proj}
\end{equation}
where $(\mathbf{R}_{t-1\to t},\mathbf{t}_{t-1\to t})$ is the ego motion between frames, $\mathbf{v}$ the object
velocity, and $\Delta t$ the frame interval. Reprojecting $\mathbf{p}_t$ places the protected region where the
object is expected to appear in frame $t$. When ego pose is unavailable, the same stage falls back to a static
prior. This stage only places candidate protected regions; the next stages decide how much precision each region
receives.

\noindent\textbf{Semantic Importance Field.}
\label{sec:field}
The green block in Fig.~\ref{fig:overview} converts projected detections into a block-level memory-importance
field. For each detection $o$ with class $c$, box scale, image position, and confidence $\mathrm{conf}$, we assign
an importance
\begin{equation}
\omega(o)=s(c)\cdot r(\text{scale},\text{pos})\cdot h(\mathrm{conf}),
\label{eq:omega}
\end{equation}
where $s(c)$ is a data-driven class-criticality weight, $r$ rewards larger and ego-lane-proximal objects, and
$h$ is monotone in confidence. The high-priority classes include pedestrians, cyclists, motorcyclists, and
vehicles; context and static background receive lower priority. The class weights are learned from the
cross-dataset bit sensitivity of each class and are floored for vulnerable road users,
\begin{equation}
s(c)=\max\!\big(\hat{s}(c),\,s_{\mathrm{VRU}}\big),\quad c\in\mathcal{C}_{\mathrm{VRU}},
\label{eq:floor}
\end{equation}
where $\hat{s}(c)$ is the measured sensitivity, $s_{\mathrm{VRU}}$ a hard floor, and
$\mathcal{C}_{\mathrm{VRU}}=\{\text{pedestrian},\text{cyclist},\text{motorcyclist}\}$. The frame is partitioned
into $B\times B$ blocks; block $(i,j)$ takes the maximum importance of the dilated detections covering it, giving
the block field $\mathbf{M}_t(i,j)$. Thresholding $\mathbf{M}_t$ yields ordered semantic tiers. In the calibrated
implementation used in Fig.~\ref{fig:overview}, these tiers separate VRU/safety-critical blocks, vehicles,
context/background, and sky. The first three come from detections and their dilated context; the sky tier is added
by the next branch.

\subsection{Scene Priors and Precision Allocation}
\noindent\textbf{Sky Prior.}
\label{sec:sky}
The sky branch in Fig.~\ref{fig:overview} handles the main low-risk region separately from the detection field. A
lightweight segmentation network produces a sky mask, and blocks inside it are assigned to the lowest-importance
sky tier with aggressive precision $k_{\mathrm{sky}}$. Ground and road are \emph{not} assigned to this aggressive
branch; they remain in the context/background tier with precision $k_{\mathrm{bg}}$, because far-range or missed
VRUs often lie on the ground plane. The output of the semantic field and sky prior is therefore a single
$B\times B$ tier map $\ell(i,j)\in\{0,\dots,L-1\}$, not just an object/background mask.

\noindent\textbf{Hierarchical Precision Allocation.}
\label{sec:alloc}
The orange block in Fig.~\ref{fig:overview} maps the multi-tier semantic map to byte precisions. The allocation is
stated per semantic tier, although calibrated tiers can share a precision value: in our fixed configuration,
VRU/safety-critical and vehicle tiers are object tiers using $k_{\mathrm{obj}}$, context/background uses
$k_{\mathrm{bg}}$, and sky uses $k_{\mathrm{sky}}$. The general per-tier precisions are solved as an
energy--distortion problem,
\begin{equation}
\min_{\{k_\ell\}}\;\sum_{\ell=0}^{L-1} a_\ell\,\rho_1(k_\ell)\quad
\text{s.t.}\quad \sum_{\ell=0}^{L-1} w_\ell\,D(k_\ell)\le D_{\max},
\label{eq:alloc}
\end{equation}
where $a_\ell$ is the area fraction of tier $\ell$, $w_\ell$ its importance weight, and $\rho_1(k)$, $D(k)$ are
the measured bit-1 density and distortion at precision $k$. Forming the Lagrangian $\mathcal{L}=\sum_\ell
a_\ell\rho_1(k_\ell)+\mu\big(\sum_\ell w_\ell D(k_\ell)-D_{\max}\big)$, the stationarity condition gives the
optimal precision of each tier,
\begin{equation}
\frac{a_\ell\,\rho_1'(k_\ell^\star)}{w_\ell\,D'(k_\ell^\star)}=-\mu\quad\forall\ell,
\label{eq:kkt}
\end{equation}
so all tiers are driven to a common marginal energy-per-distortion before any calibrated tying of tiers that share
the same operating point. Object tiers use a near-lossless precision $k_{\mathrm{obj}}$ with a per-block adaptive
mode: each block selects the mode that minimizes a joint
distortion--switching cost,
\begin{equation}
m^\star(i,j)=\arg\min_{m\in\mathcal{M}}\;d_m(i,j)+\lambda\,\alpha_m(i,j),
\label{eq:mode}
\end{equation}
where $\mathcal{M}=\{\text{preserve},\text{truncate},\text{adaptive}\}$, $d_m$ is the reconstruction distortion,
$\alpha_m$ the resulting switching, and $\lambda$ trades the two. Background tiers use the coarser
$k_{\mathrm{bg}}$, and sky blocks use $k_{\mathrm{sky}}$. The conference scheme is the special case $L=2$ with a
single shared precision and equal weights, so MotiMem-$\Omega$ strictly generalizes it.

\subsection{Byte-Level Memory Interface}
\noindent\textbf{Byte-Level Memory Coding.}
\label{sec:coding}
The blue block in Fig.~\ref{fig:overview} turns the tier decision into the actual stored bytes. For a byte $v$ at
precision $k$, the top $k$ bits $\tau=\lfloor v/2^{8-k}\rfloor$ are Gray-coded, selectively inverted when more
than half of them are set, and the inversion flag is kept in the least-significant bit:
\begin{equation}
\begin{aligned}
f&=\big[\operatorname{popc}(\operatorname{gray}(\tau))>\tfrac{k}{2}\big],\\
\operatorname{enc}_k(v)&=\big(\operatorname{gray}(\tau)\!\oplus\! f\big)\,\Vert\,\mathbf{0}\,\Vert\,f,
\end{aligned}
\label{eq:enc}
\end{equation}
where $\Vert$ denotes bit concatenation and $\mathbf{0}$ pads the dropped bits. Gray coding reduces transitions
between adjacent levels, and the selective inversion rewrites majority-one codewords to majority-zero codewords.
Thus the stored stream lowers the two quantities measured at the frame buffer: bit-1 density $\rho_1$ and
switching activity $\alpha$.

\noindent\textbf{Frame Buffer, Decode, and Energy Meter.}
\label{sec:energyrationale}
The encoded stream is written to the DRAM/SRAM frame buffer as
$\mathbf{b}=\Phi(\mathbf{I}_t;\mathcal{D}_{t-1})$, matching the gray memory block in Fig.~\ref{fig:overview}.
On read-out, $\Psi$ decodes $\mathbf{b}$ into $\tilde{\mathbf{I}}_t$, which is passed to the unmodified detection,
occupancy, and segmentation models. The detector output forms the next $\mathcal{D}_t$, which closes the
semantic-prior loop for frame $t{+}1$.

The frame-buffer meter records $\rho_1$ and $\alpha$ on the stored stream. These two quantities connect the
byte-level coder to memory-interface energy. In both DRAM and SRAM, writing and
holding a charged ($=1$) cell costs more than a discharged one; this 0/1 asymmetry is exactly what JEDEC
standardized data-bus inversion to exploit in DDR4 \cite{jedec}, and what approximate-DRAM schemes exploit by
relaxing the refresh of low-value cells \cite{raha,eden,voltron}. The selective-bit-dropping primitive we build
on measures a $39.8\%$ DRAM power reduction from this bit-density lever \cite{jetcas}. The dynamic energy of each
bus transition follows $E\!\propto\!C V^2$, so reducing $\alpha$ lowers the sensor$\rightarrow$memory and
memory$\rightarrow$processor transfer energy \cite{stanburleson}. Sec.~\ref{sec:energy} maps the measured
$(\rho_1,\alpha)$ to a derived interface-energy reduction through the coefficient-swept model.

\begin{table*}[t]
\centering
\caption{Comprehensive detection comparison: every encoding method (rows) on every metric, averaged over the 29
detectors. Rows: $\Omega$\,=\,MotiMem-$\Omega$, base\,=\,conference MotiMem~\cite{motimem}, then the naive
truncations (global-k4, uniform-4b) and the image codecs JPEG~\cite{jpeg}, WebP~\cite{webp}, PNG~\cite{png},
JPEG2000~\cite{jpeg2000}, AVIF~\cite{avif}, HEIF~\cite{heif}. Left block: mAP retention per dataset $+$ average; right blocks: aggregate all-class and VRU recall
(R), AP@IoU$0.5/0.75$ (A50/A75), COCO mAP, energy proxies $\rho_1,\alpha$ (both normalized so the original frame
is $1.00$), and SSIM/PSNR/LPIPS. All figures are
retention vs.\ the original-frame predictions ($\uparrow$; $\rho_1,\alpha$,\,LPIPS~$\downarrow$). MotiMem-$\Omega$
(bold) gives the highest average mAP among the energy-reducing methods while using low human-visible fidelity
(low PSNR), indicating that the bit budget is concentrated on task-relevant content; codecs match retention only at
$\rho_1{\approx}1$.}
\label{tab:det}
\fontsize{6.5}{6.5}\selectfont
\setlength{\tabcolsep}{0.20pt}
\renewcommand{\arraystretch}{0.9}
\begin{tabular*}{\textwidth}{@{\extracolsep{\fill}}l*{13}{c}@{\hspace{2pt}}*{3}{c}@{\hspace{2pt}}*{4}{c}@{\hspace{2pt}}*{2}{c}@{\hspace{2pt}}*{3}{c}@{}}
\toprule
& \multicolumn{13}{c}{Per-dataset mAP retention\dcite{coco}} & \multicolumn{3}{c}{All-class\dcite{coco}} & \multicolumn{4}{c}{VRU} & \multicolumn{2}{c}{Energy\dcite{stanburleson}} & \multicolumn{3}{c}{Image qual.} \\
\cmidrule(lr){2-14}\cmidrule(lr){15-17}\cmidrule(lr){18-21}\cmidrule(lr){22-23}\cmidrule(lr){24-26}
Method & nuSc\dcite{nuscenes} & A2D2\dcite{a2d2} & AV2\dcite{av2} & BDD\dcite{bdd100k} & CADC\dcite{cadc} & CaR\dcite{carina} & JODD\dcite{jodd} & KUAS\dcite{kuas} & ONCE\dcite{once} & PnSt\dcite{pandaset} & USyd\dcite{usyd} & Way\dcite{waymo} & Avg & R & A50 & A75 & R & A50 & A75 & mAP & $\rho_1$ & $\alpha$ & SSIM\dcite{ssim} & PSNR & LPIPS\dcite{lpips} \\
\midrule
Original & 1.00 & 1.00 & 1.00 & 1.00 & 1.00 & 1.00 & 1.00 & 1.00 & 1.00 & 1.00 & 1.00 & 1.00 & 1.00 & 1.00 & 1.00 & 1.00 & 1.00 & 1.00 & 1.00 & 1.00 & 1.00 & 1.00 & 1.000 & $\infty$ & .000 \\
\midrule
\textbf{$\Omega$} & \textbf{.90} & \textbf{.95} & \textbf{.89} & \textbf{.81} & \textbf{.82} & \textbf{.90} & \textbf{.94} & \textbf{.90} & \textbf{.92} & \textbf{.93} & \textbf{.91} & \textbf{.91} & \textbf{.90} & \textbf{.91} & \textbf{.92} & \textbf{.92} & \textbf{.91} & \textbf{.93} & \textbf{.93} & \textbf{.91} & \textbf{.48} & \textbf{.52} & \textbf{.953} & \textbf{25.0} & \textbf{.168} \\
base\,\cite{motimem} & .83 & .91 & .87 & .66 & .81 & .79 & .88 & .80 & .87 & .88 & .82 & .83 & .83 & .84 & .87 & .86 & .79 & .84 & .83 & .79 & .63 & .35 & .940 & 30.5 & .242 \\
global-k4 & .77 & .89 & .84 & .62 & .79 & .79 & .85 & .74 & .84 & .84 & .78 & .80 & .80 & .81 & .86 & .84 & .74 & .82 & .80 & .74 & .58 & .28 & .931 & 29.8 & .263 \\
uniform-4b & .77 & .89 & .84 & .62 & .79 & .79 & .85 & .74 & .84 & .84 & .78 & .80 & .80 & .81 & .86 & .84 & .74 & .82 & .80 & .74 & .48 & .26 & .929 & 29.2 & .264 \\
\midrule
JPEG q50 & .81 & .87 & .86 & .74 & .74 & .80 & .87 & .82 & .99 & .82 & .88 & .83 & .84 & .85 & .89 & .88 & .80 & .88 & .87 & .82 & .96 & 1.49 & .974 & 40.9 & .079 \\
JPEG q100 & .98 & .98 & .98 & .96 & .96 & .96 & .98 & .99 & .98 & .98 & .98 & .98 & .98 & .98 & .98 & .98 & .97 & .98 & .98 & .98 & 1.24 & 1.54 & 1.000 & 57.2 & .001 \\
WebP q50 & .76 & .84 & .85 & .70 & .72 & .74 & .84 & .76 & .83 & .83 & .83 & .79 & .79 & .81 & .86 & .84 & .72 & .82 & .80 & .74 & 1.03 & 1.54 & .955 & 38.7 & .139 \\
WebP q100 & .95 & .97 & .97 & .92 & .94 & .95 & .97 & .97 & .96 & .96 & .96 & .96 & .96 & .96 & .97 & .96 & .95 & .97 & .96 & .96 & 1.03 & 1.54 & .995 & 48.4 & .012 \\
JPEG2000 & .98 & .97 & .99 & .96 & .98 & .99 & .98 & .95 & .98 & .98 & .99 & .98 & .98 & .98 & .98 & .98 & .97 & .98 & .98 & .98 & 1.03 & 1.54 & .998 & 56.1 & .002 \\
PNG & 1.00 & 1.00 & 1.00 & 1.00 & 1.00 & 1.00 & 1.00 & 1.00 & 1.00 & 1.00 & 1.00 & 1.00 & 1.00 & 1.00 & 1.00 & 1.00 & 1.00 & 1.00 & 1.00 & 1.00 & 1.03 & 1.53 & 1.000 & $\infty$ & .000 \\
AVIF q50 & .77 & .84 & .86 & .71 & .71 & .76 & .85 & .77 & .86 & .83 & .84 & .80 & .80 & .82 & .87 & .85 & .74 & .83 & .81 & .76 & 1.02 & 1.53 & .961 & 39.3 & .148 \\
HEIF q50 & .88 & .91 & .93 & .84 & .85 & .87 & .91 & .89 & .90 & .90 & .91 & .90 & .89 & .90 & .93 & .92 & .86 & .91 & .91 & .87 & 1.03 & 1.54 & .989 & 44.3 & .029 \\
\bottomrule
\end{tabular*}
\end{table*}
\section{Experiments}
\subsection{Setup and Implementation Details}
\begin{table}[!ht]
\centering
\caption{Per-dataset detection retention and energy for the three key methods (12 datasets, each averaged over
29 detectors). base\,=\,MotiMem~\cite{motimem}, nv\,=\,naive global-k4; R\,=\,recall, A50/A75\,=\,AP@IoU$0.5/0.75$
(middle block: VRU); $\rho_1$/$\alpha$ are the energy proxies, both normalized to the original frame
(per-dataset values scaled to the Table~\ref{tab:energy} means). The 12 sets span five continents and
day/night/snow/desert. $\Omega$ (bold) has the strongest overall retention profile among the energy-reducing
methods and the lowest $\rho_1$, with one near-tie (CADC A75, $.85$ vs $.86$).}
\label{tab:detperds}
\footnotesize
\setlength{\tabcolsep}{1.2pt}
\renewcommand{\arraystretch}{0.9}
\begin{tabular}{@{}lllcccccccccc@{}}
\toprule
& & & \multicolumn{4}{c}{All classes} & \multicolumn{4}{c}{VRU} & \multicolumn{2}{c}{Energy} \\
\cmidrule(lr){4-7}\cmidrule(lr){8-11}\cmidrule(lr){12-13}
Dataset & Location & & R & A50 & A75 & mAP & R & A50 & A75 & mAP & $\rho_1$ & $\alpha$ \\
\midrule
\multirow{3}{*}{nuSc\,\cite{nuscenes}} & \multirow{3}{*}{US, Boston} & $\Omega$ & \textbf{.92} & \textbf{.92} & \textbf{.92} & \textbf{.90} & \textbf{.91} & \textbf{.95} & \textbf{.94} & \textbf{.93} & .57 & .50 \\
 & & base & .84 & .88 & .86 & .83 & .76 & .82 & .81 & .77 & .63 & .38 \\
 & & nv & .79 & .84 & .82 & .77 & .70 & .80 & .79 & .73 & .58 & .29 \\
\midrule
\multirow{3}{*}{A2D2\,\cite{a2d2}} & \multirow{3}{*}{Germany} & $\Omega$ & \textbf{.96} & \textbf{.96} & \textbf{.96} & \textbf{.95} & \textbf{.93} & \textbf{.95} & \textbf{.95} & \textbf{.94} & .46 & .51 \\
 & & base & .92 & .94 & .93 & .91 & .85 & .90 & .89 & .87 & .63 & .38 \\
 & & nv & .90 & .93 & .92 & .89 & .78 & .87 & .86 & .82 & .59 & .32 \\
\midrule
\multirow{3}{*}{AV2\,\cite{av2}} & \multirow{3}{*}{US, 6 cities} & $\Omega$ & \textbf{.90} & \textbf{.92} & \textbf{.91} & \textbf{.89} & \textbf{.88} & \textbf{.88} & \textbf{.88} & \textbf{.85} & .47 & .51 \\
 & & base & .88 & .90 & .90 & .87 & .70 & .66 & .64 & .61 & .66 & .41 \\
 & & nv & .85 & .89 & .88 & .84 & .64 & .64 & .61 & .57 & .59 & .32 \\
\midrule
\multirow{3}{*}{BDD\,\cite{bdd100k}} & \multirow{3}{*}{US, NY/SF} & $\Omega$ & \textbf{.82} & \textbf{.84} & \textbf{.83} & \textbf{.81} & \textbf{.85} & \textbf{.88} & \textbf{.87} & \textbf{.86} & .41 & .51 \\
 & & base & .69 & .72 & .69 & .66 & .72 & .77 & .75 & .73 & .53 & .31 \\
 & & nv & .65 & .69 & .66 & .62 & .64 & .69 & .67 & .63 & .48 & .24 \\
\midrule
\multirow{3}{*}{CADC\,\cite{cadc}} & \multirow{3}{*}{Canada} & $\Omega$ & \textbf{.85} & \textbf{.88} & \textbf{.85} & \textbf{.82} & \textbf{.93} & \textbf{.96} & \textbf{.96} & \textbf{.94} & .40 & .38 \\
 & & base & .83 & .87 & .86 & .81 & .82 & .91 & .90 & .85 & .59 & .28 \\
 & & nv & .81 & .86 & .84 & .79 & .79 & .91 & .90 & .83 & .55 & .22 \\
\midrule
\multirow{3}{*}{CaR\,\cite{carina}} & \multirow{3}{*}{Brazil} & $\Omega$ & \textbf{.92} & \textbf{.92} & \textbf{.92} & \textbf{.90} & \textbf{.89} & \textbf{.92} & \textbf{.91} & \textbf{.90} & .75 & .57 \\
 & & base & .81 & .84 & .83 & .79 & .78 & .87 & .85 & .82 & .85 & .33 \\
 & & nv & .81 & .86 & .84 & .79 & .64 & .79 & .76 & .69 & .78 & .27 \\
\midrule
\multirow{3}{*}{JODD\,\cite{jodd}} & \multirow{3}{*}{Japan} & $\Omega$ & \textbf{.94} & \textbf{.95} & \textbf{.95} & \textbf{.94} & \textbf{.94} & \textbf{.96} & \textbf{.96} & \textbf{.95} & .45 & .57 \\
 & & base & .89 & .91 & .90 & .88 & .87 & .91 & .90 & .86 & .62 & .38 \\
 & & nv & .87 & .90 & .88 & .85 & .83 & .90 & .88 & .84 & .57 & .30 \\
\midrule
\multirow{3}{*}{KUAS\,\cite{kuas}} & \multirow{3}{*}{UAE} & $\Omega$ & \textbf{.91} & \textbf{.92} & \textbf{.92} & \textbf{.90} & \textbf{.92} & \textbf{.94} & \textbf{.94} & \textbf{.93} & .42 & .33 \\
 & & base & .81 & .84 & .83 & .80 & .82 & .87 & .86 & .83 & .57 & .19 \\
 & & nv & .76 & .81 & .79 & .74 & .74 & .82 & .80 & .74 & .54 & .15 \\
\midrule
\multirow{3}{*}{ONCE\,\cite{once}} & \multirow{3}{*}{China} & $\Omega$ & \textbf{.93} & \textbf{.94} & \textbf{.94} & \textbf{.92} & \textbf{.89} & \textbf{.90} & \textbf{.90} & \textbf{.87} & .43 & .56 \\
 & & base & .88 & .91 & .90 & .87 & .76 & .79 & .77 & .71 & .58 & .39 \\
 & & nv & .85 & .89 & .88 & .84 & .71 & .77 & .75 & .68 & .53 & .31 \\
\midrule
\multirow{3}{*}{PnSt\,\cite{pandaset}} & \multirow{3}{*}{US, SF} & $\Omega$ & \textbf{.94} & \textbf{.95} & \textbf{.95} & \textbf{.93} & \textbf{.93} & \textbf{.95} & \textbf{.95} & \textbf{.93} & .44 & .54 \\
 & & base & .89 & .92 & .91 & .88 & .83 & .88 & .87 & .82 & .62 & .36 \\
 & & nv & .86 & .90 & .89 & .84 & .79 & .86 & .84 & .77 & .56 & .27 \\
\midrule
\multirow{3}{*}{USyd\,\cite{usyd}} & \multirow{3}{*}{Australia} & $\Omega$ & \textbf{.92} & \textbf{.94} & \textbf{.93} & \textbf{.91} & \textbf{.92} & \textbf{.95} & \textbf{.94} & \textbf{.92} & .53 & .60 \\
 & & base & .84 & .87 & .86 & .82 & .82 & .89 & .87 & .83 & .67 & .39 \\
 & & nv & .80 & .85 & .83 & .78 & .79 & .88 & .86 & .80 & .62 & .31 \\
\midrule
\multirow{3}{*}{Way\,\cite{waymo}} & \multirow{3}{*}{US, SF/Phx} & $\Omega$ & \textbf{.92} & \textbf{.94} & \textbf{.93} & \textbf{.91} & \textbf{.91} & \textbf{.95} & \textbf{.95} & \textbf{.93} & .44 & .60 \\
 & & base & .84 & .88 & .87 & .83 & .78 & .88 & .86 & .82 & .62 & .41 \\
 & & nv & .82 & .87 & .85 & .80 & .76 & .87 & .86 & .80 & .57 & .33 \\
\bottomrule
\end{tabular}
\end{table}
\textbf{Tasks and models.} We evaluate three perception tasks. Detection uses 29 detectors over 12 driving
datasets; 3D occupancy uses the 5 models of Table~\ref{tab:occ} on nuScenes; semantic segmentation uses the 5
architectures of Table~\ref{tab:seg}, with DeepLabV3 spanning all 12 datasets. \textbf{Metric.} We report
\emph{retention}: each model is run on the decoded frame and on the original frame, and we measure how well the
two predictions agree (mAP and VRU recall for detection, mIoU/RayIoU for occupancy, mIoU for segmentation). The
energy axis is the bit-1 density $\rho_1$ and switching activity $\alpha$ of Eq.~(1), normalized so the
original frame is $1.0$. \textbf{Baselines.} The conference scheme (\emph{base}, MotiMem~\cite{motimem}, single $k{=}4$); two
energy-matched naive truncations (\emph{global-k4}, \emph{uniform-4b}); and six image codecs (JPEG, WebP, PNG,
JPEG2000, AVIF, HEIF). \textbf{Compute.} Detection, segmentation, and all
$\Omega$ encoding run on a single NVIDIA RTX PRO 6000 (Blackwell) GPU server, with the larger occupancy models
on NVIDIA A100 GPUs. The $\Omega$ encoder itself is a lightweight numpy/OpenCV transform applied
to each frame before the model, with no dedicated accelerator.

\subsection{Bit-Budget Calibration}
\label{sec:calib}
\begin{figure*}[t]
\centering
\includegraphics[width=\textwidth]{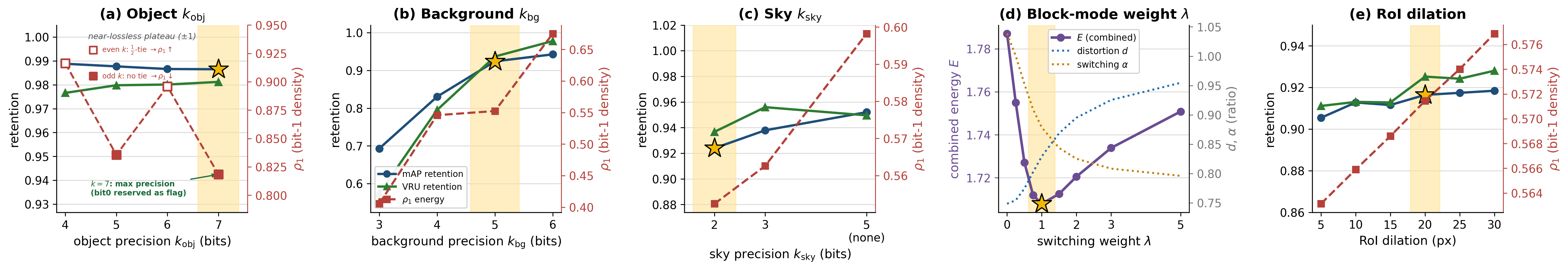}
\caption{Cross-dataset operating-point sweep used to choose the fixed configuration ($12$ datasets $\times$ $29$
detectors). Each panel varies one design variable and plots retention against the energy proxies; stars mark the
selected points. (a)~Object precision $k_{\mathrm{obj}}$: all usable values $k{\le}7$ are near-lossless
($\pm1$-bit flip, flat ${\sim}98\%$ retention), and $k{=}7$ is both the maximum precision and an odd-$k$
low-$\rho_1$ point. (b)~Background precision $k_{\mathrm{bg}}$: retention saturates after $k{=}5$ and $\rho_1$
jumps at $k{=}6$, so $k_{\mathrm{bg}}{=}5$. (c)~Sky precision $k_{\mathrm{sky}}$: sky moves from untruncated
$k{=}5$ to $k_{\mathrm{sky}}{=}2$, returning energy at a small accuracy cost. (d)~Block-mode weight $\lambda$:
the combined energy is minimized at $\lambda{=}1$. (e)~RoI dilation: VRU recall saturates at $20$ px.}
\label{fig:calib}
\end{figure*}

Fig.~\ref{fig:calib} is the calibration experiment behind the fixed $k_{\mathrm{obj}}$, $k_{\mathrm{bg}}$, and
$k_{\mathrm{sky}}$ used in all later tables. We sweep one design variable at a time on the 12-dataset,
29-detector validation set and record both task retention and the energy proxies. The selected point is not the
most accurate point alone; it is the knee that returns the largest bit-density or switching reduction before
retention begins to fall sharply.
\textbf{(a)~Object precision.} Objects carry the safety-critical content, so they are coded near-lossless: the
block-adaptive coder flips only the LSB (reconstruction error $\pm1$) for every $k_{\mathrm{obj}}{<}8$, so object
retention stays on a flat ${\sim}98\%$ near-lossless plateau (both mAP and VRU) across the whole sweep. Bit~$0$ is
permanently the flip flag the decoder reads to undo the inversion, so $k_{\mathrm{obj}}{=}8$ would leave no flag
bit and does not exist; the usable range is $k_{\mathrm{obj}}{\le}7$ and $k_{\mathrm{obj}}{=}7$ is the maximum
precision. The energy proxy $\rho_1$ is not monotonic in $k$ but alternates with its parity, because the
bus-invert flip inverts a $k$-bit codeword only when it carries more than $k/2$ ones. For even $k$ ($4,6$) the
codewords with exactly $k/2$ ones are ties the flip cannot improve, so they stay at density $0.5$ and raise
$\rho_1$ (e.g.\ $k{=}6$ sits at $\rho_1{\approx}0.90$); for odd $k$ ($5,7$) no tie exists and every codeword is
driven below $k/2$, so $\rho_1$ falls to ${\approx}0.82$. The chosen $k_{\mathrm{obj}}{=}7$ is therefore doubly
optimal: the highest usable precision, and an odd level with the lowest $\rho_1$. \textbf{(b)~Background precision.} mAP and VRU retention rise steeply up to
$k_{\mathrm{bg}}{=}5$ ($+9$ pp mAP over $k_{\mathrm{bg}}{=}4$ at essentially the same $\rho_1{\approx}0.55$) and
then saturate, whereas $k_{\mathrm{bg}}{=}6$ buys only $+2$ pp mAP for a $22\%$ jump in $\rho_1$; we fix
$k_{\mathrm{bg}}{=}5$ at the knee. \textbf{(c)~Sky precision.} Sky is empty for every perception task, so it is
truncated aggressively: dropping the sky tier from its untruncated $k{=}5$ to $k_{\mathrm{sky}}{=}2$ cuts $\rho_1$
from $0.60$ to $0.55$ for about $3$ pp mAP and $1$ pp VRU recall, the cheapest energy in the frame.
\textbf{(d)~Block-mode weight.} Each block minimizes
$d+\lambda\alpha$, trading reconstruction distortion $d$ against switching activity $\alpha$; the combined energy
is minimized at $\lambda{=}1$. \textbf{(e)~RoI dilation.} VRU recall saturates at $20$ px and wider dilation only
raises $\rho_1$, so we dilate by $20$ px. These sweeps show directly how the precision choices affect performance:
object precision is accuracy-limited, background precision is knee-limited, and sky precision is energy-limited.
None of the operating points is tuned to a single domain. The resulting calibrated configuration, fixed across
all models and datasets in the later tables, is: object tiers $k_{\mathrm{obj}}{=}7$ with block-adaptive
multi-mode, background $k_{\mathrm{bg}}{=}5$, sky $k_{\mathrm{sky}}{=}2$, ground kept at $k_{\mathrm{bg}}$,
$\lambda{=}1$, RoI dilation $20$ px, and $L{=}4$ tiers.

\subsection{Detection}

\begin{figure}[!ht]
\centering
\includegraphics[width=\columnwidth]{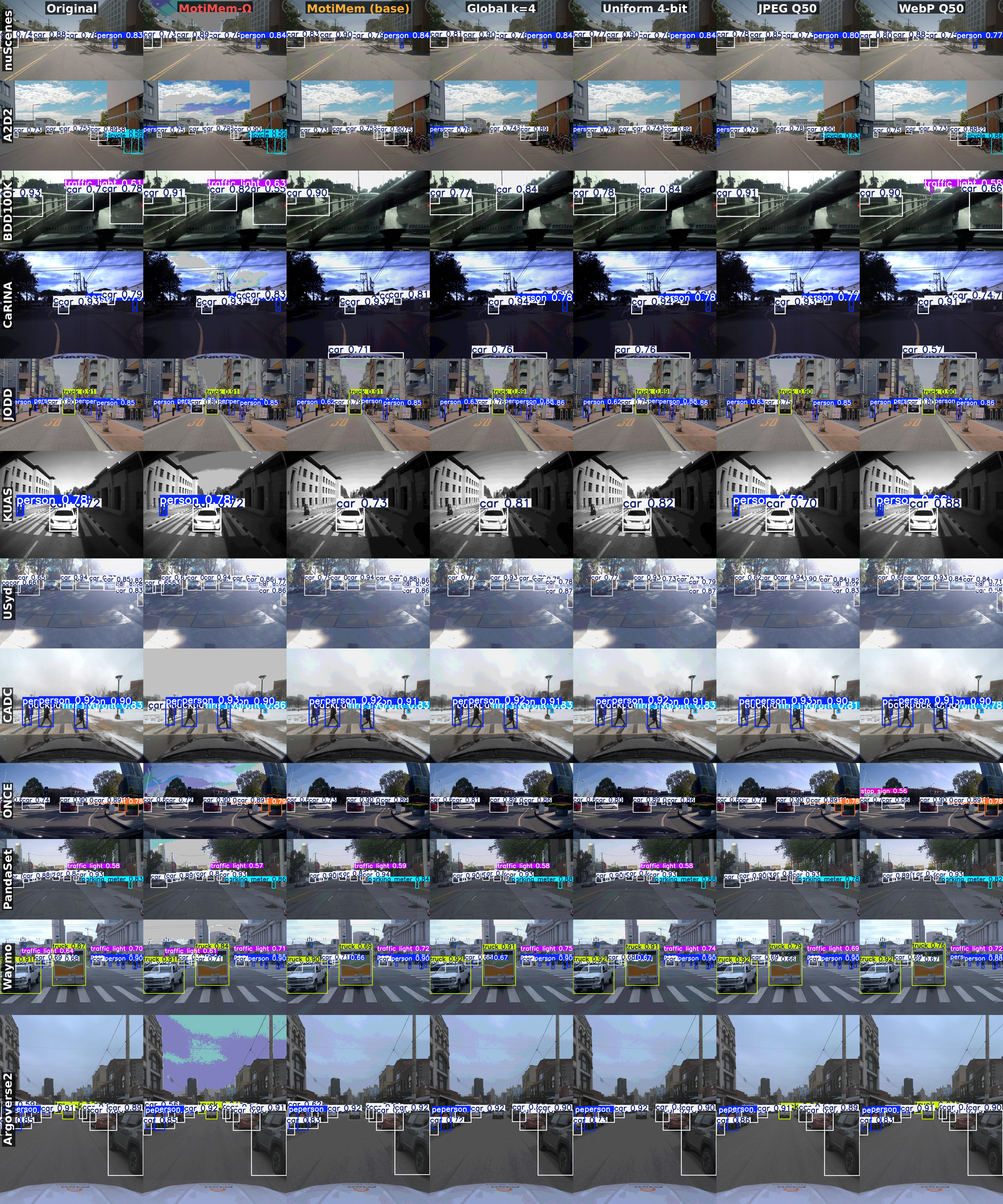}
\caption{Qualitative detection over the 12 datasets (rows) $\times$ seven encoding methods (columns), with
native YOLO26 boxes on each decoded frame. \mbox{MotiMem-$\Omega$} (red) keeps detections that the conference
base \mbox{MotiMem}~\cite{motimem} (orange) and the energy-matched naive truncations (global-k4, uniform-4b)
drop: the base has no motion-compensated closed loop, so an object that moved out of its $t{-}1$ RoI is
truncated and lost, whereas $\Omega$'s tiered allocation keeps it. Image codecs alter the bitstream but not the
predictions. Snow (CADC), night (BDD), and clear scenes are covered.}
\label{fig:det_qual}
\end{figure}

Table~\ref{tab:det} reports every method on every metric and every dataset, and Figure~\ref{fig:det_qual}
shows the comparison qualitatively. \mbox{MotiMem-$\Omega$} retains
$0.90$ aggregate mAP and $0.91$ VRU recall, compared with $0.83$/$0.79$ for the baseline, while also using
lower bit-1 density. Across the 29 detectors, the per-model mAP deltas over the baseline are $+0.05$ to $+0.10$, and
the 12-dataset average is higher than the energy-matched naive truncations ($0.80$ mAP) at comparable density.
This indicates that the saving comes from \emph{where} the bits are dropped, not only how many. The image-quality
columns make the point sharply: \mbox{MotiMem-$\Omega$} has low PSNR ($25.0$\,dB) but high task retention, because
it spends bits on objects rather than on human-visible background fidelity, whereas the codecs reach
$40$--$57$\,dB PSNR but save no interface energy. Table~\ref{tab:detperds} shows the same trend metric by metric,
with one near-tie (CADC AP@$0.75$, $.85$ vs the baseline's $.86$).

\subsection{Semantic Segmentation}

\begin{table}[!ht]
\centering
\caption{Segmentation mIoU retention (vs.\ original-frame map) under every encoding method, on five architectures
(nuScenes; base\,=\,MotiMem~\cite{motimem}, g-k4\,=\,global-$k$4, unif\,=\,uniform-4bit; $\rho_1$ row\,=\,bit-1
density). $\Omega$ (bold) leads on every architecture; JPEG only matches it at $\rho_1{\approx}1$ (no saving).
The lead also holds across 12 datasets (DeepLabV3 $0.906$ vs.\ $0.858$, $+4.8$ pp; range $+1.2$ to $+10.7$).}
\label{tab:seg}
\small
\setlength{\tabcolsep}{4.5pt}
\begin{tabular*}{\columnwidth}{@{\extracolsep{\fill}}lccccc@{}}
\toprule
Architecture & $\Omega$ & base & g-k4 & unif & JPEG \\
\textit{$\rho_1$} & \textit{.57} & \textit{.63} & \textit{.58} & \textit{.48} & \textit{.96} \\
\midrule
DeepLabV3~\cite{deeplabv3} & \textbf{.934} & .899 & .885 & .886 & .904 \\
FCN~\cite{fcn} & \textbf{.889} & .861 & .827 & .827 & .862 \\
LRASPP~\cite{lraspp} & \textbf{.836} & .789 & .769 & .770 & .817 \\
SegFormer~\cite{segformer} & \textbf{.715} & .626 & .625 & .625 & .713 \\
Mask2Former~\cite{mask2former} & \textbf{.758} & .702 & .701 & .701 & .754 \\
\bottomrule
\end{tabular*}
\end{table}

\begin{figure}[!ht]
\centering
\includegraphics[width=\columnwidth]{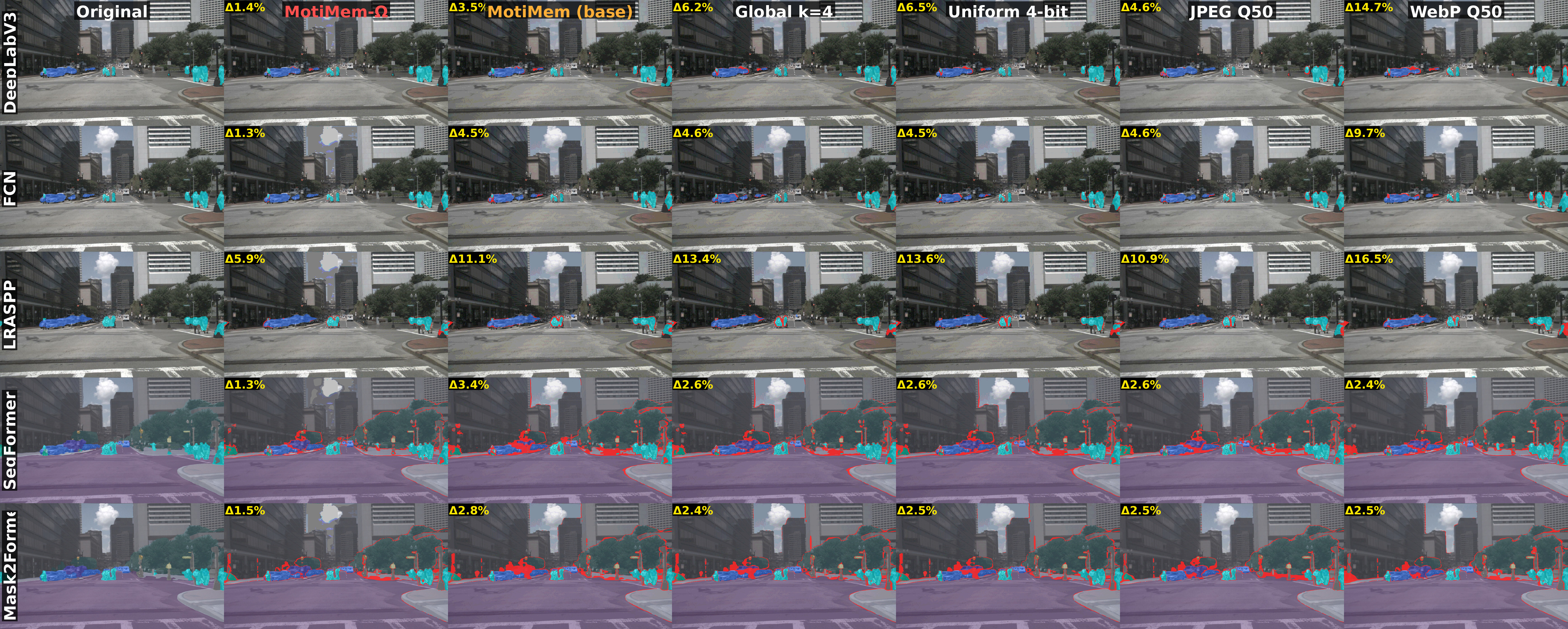}
\caption{Semantic segmentation under the seven encoding methods (columns) across the five architectures of
Table~\ref{tab:seg} (rows; DeepLabV3 / FCN / LRASPP are Pascal-VOC, SegFormer / Mask2Former are Cityscapes), on
one nuScenes frame. Each cell blends the model's seg map over the decoded frame; mislabelled pixels (class changed
vs.\ the original-frame map) are highlighted in \emph{red}, with the error-\% (excluding the deliberately-truncated
sky/background) printed. The decoded frames are near-identical by design (MotiMem preserves visual quality), so the
visible signal is the downstream seg-map error, not the pixels. \mbox{MotiMem-$\Omega$} (the second column)
reproduces the original labelling with almost no error across all five models.}
\label{fig:seg_qual}
\end{figure}
Table~\ref{tab:seg} (qualitatively Fig.~\ref{fig:seg_qual}) shows that \mbox{MotiMem-$\Omega$} preserves more
segmentation than the energy-matched naive truncation on all five architectures, by $+4.9$ to $+9.0$ pp, and the advantage holds across the 12 datasets
(DeepLabV3 mean $+4.8$ pp, per-dataset range $+1.2$ to $+10.7$ pp). Over all $23$ (architecture, dataset)
cells, $\Omega$ beats naive in $22$, the single near-tie being SegFormer on A2D2 ($-1.4$ pp on the
lowest-gap set). Together with occupancy, this establishes dense perception as the regime where semantic
allocation is decisive.

\subsection{3D Occupancy}

\begin{table}[!ht]
\centering
\caption{3D occupancy retention (\% of each model's native mIoU/RayIoU vs.\ original) under every encoding method,
on five SOTA models (base\,=\,MotiMem~\cite{motimem}, g-k4\,=\,global-$k$4, unif\,=\,uniform-4bit; $\rho_1$
row\,=\,bit-1 density). $\Omega$ (bold) retains the most among energy-saving methods; base and naive lose $2$--$7$
pp, while JPEG matches only at $\rho_1{>}1$ (no saving).}
\label{tab:occ}
\small
\setlength{\tabcolsep}{4pt}
\begin{tabular}{@{}llccccc@{}}
\toprule
Model & Metric & $\Omega$ & base & g-k4 & unif & JPEG \\
& \textit{$\rho_1$} & \textit{.59} & \textit{.63} & \textit{.59} & \textit{.45} & \textit{1.04} \\
\midrule
GaussRender~\cite{gaussrender} & mIoU & \textbf{99.7} & 94.8 & 92.8 & 92.8 & 97.7 \\
DAOcc~\cite{daocc} & mIoU & \textbf{99.9} & 97.8 & 97.5 & 97.5 & 99.6 \\
OPUS~\cite{opus} & mIoU & \textbf{99.6} & 96.8 & 95.7 & 95.6 & 99.1 \\
GaussianFormer-2~\cite{gaussianformer2} & mIoU & \textbf{98.2} & 96.9 & 96.3 & 96.4 & 99.2 \\
SparseOcc~\cite{sparseocc} & RayIoU & \textbf{99.8} & 96.3 & 95.6 & 95.5 & 99.1 \\
\bottomrule
\end{tabular}
\end{table}

\begin{figure}[!ht]
\centering
\includegraphics[width=\columnwidth]{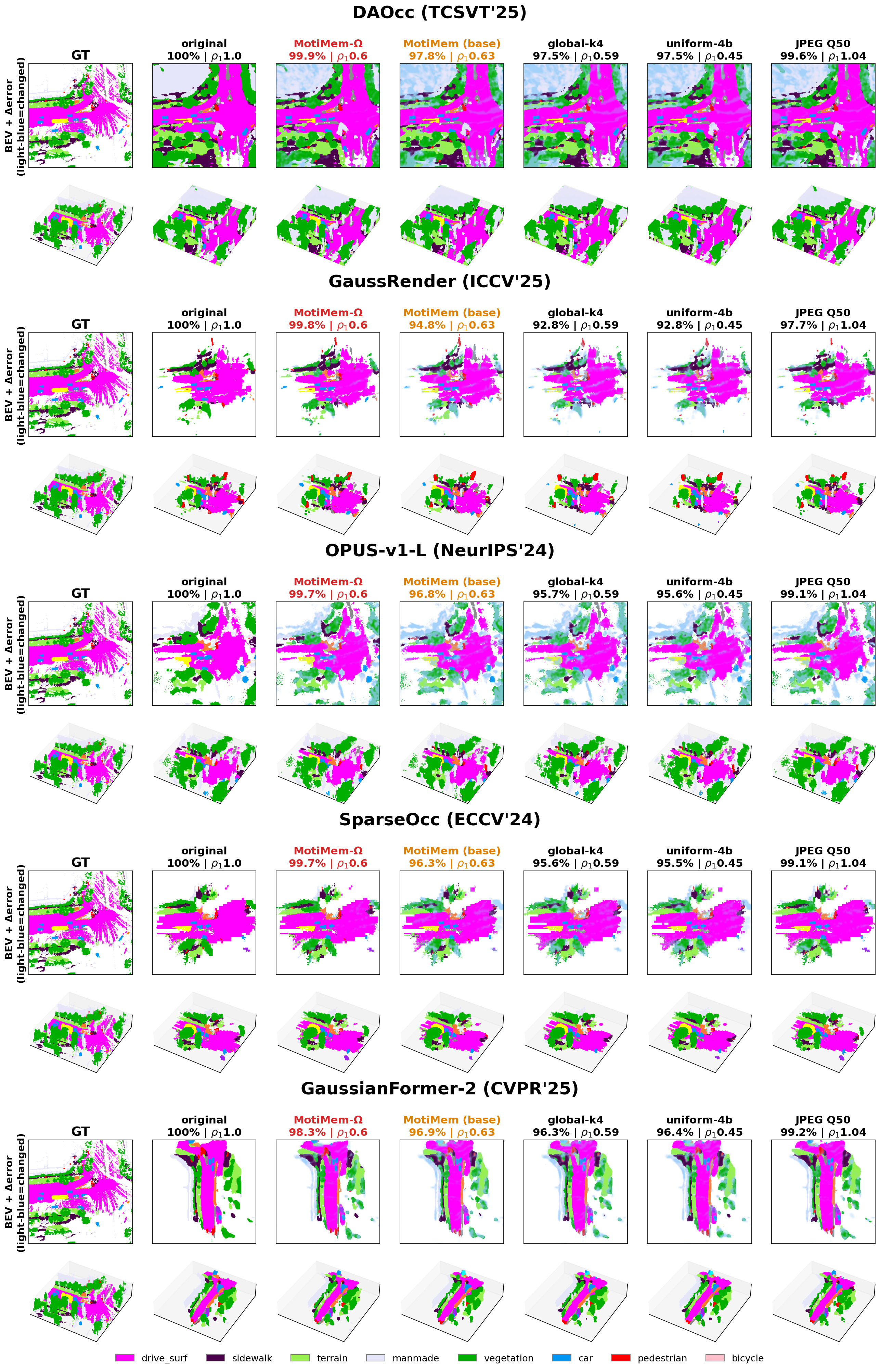}
\caption{3D occupancy on the five SOTA models (rows): BEV (top) $+$ 3D (bottom) under the original input vs.\
\mbox{MotiMem-$\Omega$} (red), the conference base \mbox{MotiMem}~\cite{motimem} (orange), the non-semantic
truncations, and JPEG; changed voxels are highlighted in light blue. Each column title gives the retention
(\% of the model's native mIoU/RayIoU) and the bit-1 density $\rho_1$. $\Omega$ alone is both high-retention ($98.2$--$99.9\%$) and
low-energy ($\rho_1{=}0.6$), whereas JPEG matches retention but at $\rho_1{>}1$ (no memory saving).}

\label{fig:occ_qual}
\end{figure}
On the five state-of-the-art occupancy models (Table~\ref{tab:occ}, Fig.~\ref{fig:occ_qual}), \mbox{MotiMem-$\Omega$} retains
$98.2$--$99.9\%$ of the native score at $\rho_1{=}0.59$, while the conference base retains only $94.8$--$97.8\%$.
The gap is widest on the geometry-sensitive GaussRender ($99.7\%$ vs.\ $94.8\%$), confirming that dense
per-voxel perception is where uniform truncation hurts and where semantic allocation pays off.

\textbf{Cross-task analysis.} The energy-matched advantage over naive truncation grows with the density of the
perception output. Sparse, agent-level detection is forgiving: the detector reasons over whole objects and
tolerates a posterized background, so even the naive baselines keep most of the mAP, although $\Omega$ still leads
by $3$ to $19$ pp per dataset and leads most on the small, far VRU classes. The gap opens up on dense per-pixel and per-voxel
perception (segmentation $+4.9$ to $+9.0$ pp; occupancy $99.8\%$ vs.\ $95\%$), because these models read every
location, so a uniform truncation that flattens the background rewrites the dense label or voxel field. The same
effect explains $\Omega$'s low PSNR at high accuracy (Table~\ref{tab:det}): the truncated background that no model
reads is energy spent for nothing.

\begin{table}[!ht]
\centering
\caption{Derived memory-interface energy. From the measured $\rho_1$ and $\alpha$ ratios, the model of
Eq.~\eqref{eq:energy} yields a reduction $\Delta E$ with a coefficient-agnostic range.}
\label{tab:energy}
\small
\setlength{\tabcolsep}{5.5pt}
\begin{tabular}{@{}llccc@{}}
\toprule
Task & Method & $\rho_1$ & $\alpha$ & $\Delta E$ (\%) \\
\midrule
\multirow{3}{*}{Det/Seg}
 & \textbf{MotiMem-$\Omega$} & 0.480 & 0.516 & \textbf{35.5 [27.7, 43.3]} \\
 & naive global-k4 & 0.580 & 0.277 & 37.0 [26.1, 46.3] \\
 & JPEG q50 & 0.957 & 1.493 & $-10.4$ [$-16.0$, $-2.3$] \\
\midrule
\multirow{3}{*}{Occ}
 & \textbf{MotiMem-$\Omega$} & 0.593 & 0.526 & \textbf{30.2 [23.1, 37.0]} \\
 & naive global-k4 & 0.589 & 0.247 & 37.3 [26.0, 46.9] \\
 & JPEG q50 & 1.041 & 1.508 & $-14.6$ [$-19.9$, $-6.9$] \\
\bottomrule
\end{tabular}
\end{table}

\subsection{Derived Memory-Interface Energy}
\label{sec:energy}
Table~\ref{tab:energy} maps the measured proxies to a derived interface-energy reduction through
Eq.~\eqref{eq:energy}, sweeping the coefficients over the literature ranges ($w_0\!\in\![0.15,0.45]$,
$w_1\!\in\![0.30,0.60]$, $w_\alpha\!\in\![0.10,0.35]$). \mbox{MotiMem-$\Omega$} reduces front-camera bit-1 density
by $52\%$ and switching by $48\%$, which the model maps to a $35.5\%$ interface-energy reduction for detection
and segmentation, with a lower end of $27.7\%$ across the swept literature coefficients, and $30.2\%$ (lower end
$23.1\%$) for six-camera occupancy. The proxy is reported on each task's
own evaluation setup: detection and segmentation encode the \emph{front} camera (detection $\rho_1{=}0.48$ is the
12-dataset mean), while occupancy encodes all \emph{six surround} cameras on nuScenes ($\rho_1{=}0.59$, a denser
content), so its reduction is naturally smaller. On nuScenes alone the front-camera proxy is $\rho_1{\approx}0.57$
for both detection and segmentation, so the per-setup measurements are consistent. The naive truncations reach a slightly
higher number, but only by destroying accuracy (Tables~\ref{tab:det}--\ref{tab:seg}); $\Omega$'s value is the
joint operating point, not the raw energy. JPEG yields a \emph{negative} reduction because entropy-coded bytes
raise switching, confirming that compression is not a memory-interface-energy lever.

\subsection{Design Sensitivity Analysis}

\begin{table}[!ht]
\centering
\caption{Detection design sensitivity. Top rows perturb energy-related design choices; last rows swap the temporal
prior. Metrics are normalized to the original frame.}
\label{tab:sensitivity}
\small
\begin{tabular*}{\columnwidth}{@{\extracolsep{\fill}}lcccc@{}}
\toprule
Variant & mAP & VRU & $\rho_1$ & $\alpha$ \\
\midrule
\textbf{MotiMem-$\Omega$ (full)} & 0.900 & 0.910 & \textbf{0.480} & \textbf{0.516} \\
\;\;no Gray & 0.904 & 0.916 & 0.517 & 0.522 \\
\;\;sky at bg-$k$ & 0.932 & 0.921 & 0.514 & 0.636 \\
\;\;ego-CV & 0.903 & 0.915 & 0.482 & 0.547 \\
\;\;2D-vel. & 0.903 & 0.915 & 0.480 & 0.544 \\
\;\;static & 0.903 & 0.915 & 0.482 & 0.547 \\
\bottomrule
\end{tabular*}
\end{table}

Table~\ref{tab:sensitivity} is an operating-point sensitivity check. Removing Gray mainly costs energy
($\rho_1$: $0.480{\to}0.517$); keeping sky at background precision recovers $+3.2$ pp mAP but costs $+7.1\%$
bit-1 density and $+23.3\%$ switching, so the full model selects the lower-energy point. The temporal-prior rows
differ by at most $0.003$ mAP, showing that the gain comes from semantic allocation plus sky/byte-level coding
rather than a tuned motion model.

\section{Discussion and Conclusion}
\mbox{MotiMem-$\Omega$} changes what the memory stores
rather than what the model computes, and it is tuned to what a perception model reads rather than to
human-perceived image quality. A single encoded frame therefore serves $39$ heterogeneous models ($29$
detectors, $5$ occupancy and $5$ segmentation networks) on $12$ datasets with no per-model tuning, because all of
them discard the same content: the flat sky and texture-less road that uniform truncation and image codecs still
pay memory energy to reconstruct. Moving the bit budget off that content and onto the objects and vulnerable road
users that decisions depend on costs almost no accuracy, yet it removes most of the stored $1$-bits and bus
transitions that set the energy.

The common efficiency levers act elsewhere. Quantization and pruning shrink the
\emph{model}; Video-Coding-for-Machines and JPEG~AI shrink the \emph{transported file}. Neither lowers the bit-1
density or switching of the frame while it sits in and moves through memory, which is why our codec baselines
keep accuracy but save nothing at $\rho_1{\ge}1$. Content-blind approximate memory does lower that energy, but it
cannot separate objects from background and corrupts them. \mbox{MotiMem-$\Omega$} instead makes the
approximate-memory decision semantic, and it contains the conference scheme as the $L{=}2$, equal-importance
special case~\cite{motimem}, extending it to $L$ data-driven tiers with cross-dataset class-sensitivity weights and
a VRU safety floor. The proxies are not ad hoc either: $\rho_1$ and $\alpha$ follow from bus-encoding
theory~\cite{stanburleson}, JEDEC data-bus inversion~\cite{jedec}, and the measured DRAM-power result
of~\cite{jetcas}, and the coefficient-swept model turns the $52\%$ and $48\%$ proxy reductions into a derived
$36\%$ energy reduction, bounded below by $27\%$ for any coefficient choice, which stands in for a silicon
measurement in this submission.

Beyond these findings, three practical considerations frame deployment. In implementation terms,
\mbox{MotiMem-$\Omega$} is a drop-in transform on the camera-to-frame-buffer path: the field construction and
per-block coding are lightweight integer operations that can live in the image signal processor or in a small
logic block in front of the memory controller. They reuse the detections the perception stack already computes
and require no retraining of downstream models, so the integration cost is confined to the sensing pipeline. At
the vehicle level, the multi-camera sensing path is an always-on load, so a roughly $30$--$36\%$ cut in
memory-interface energy returns budget to computing or driving range on electric platforms and eases the thermal
envelope of camera modules, while the VRU safety floor keeps the protection of pedestrians and cyclists explicit
rather than incidental. At the transportation-system level, the per-vehicle saving scales across fleets of
camera-heavy automated vehicles and roadside perception units, and a shared semantic-importance convention for
the sensing path would let vehicles and infrastructure adopt the same energy-aware storage interface.

Although \mbox{MotiMem-$\Omega$} delivers consistent interface-energy savings across detectors,
dense-perception tasks, and datasets while holding perception accuracy, four limitations exist.
\emph{(i)~Proxy, not watts.} The energy figure is an analytical derivation:
Eq.~\eqref{eq:energy} applied to the measured bit-1 density and switching, with the coefficients swept over the
memory-energy literature (Sec.~\ref{sec:energy}), rather than a direct silicon watt measurement.
\emph{(ii)~Scope.} $\Omega$ saves the perception input data path, the camera frame buffer and the buses it
crosses, not total-vehicle, NN-compute, or actuation energy. \emph{(iii)~Content dependence.} The saving grows
with the empty-region fraction, above all the visible sky, so scenes that show little sky save less: on the
urban canyons of CaRINA $\rho_1$ stays at $0.75$ against the $0.48$ average. Ground and road are
left untruncated on purpose, as a VRU safety floor: a ground-$k$ sweep showed that any ground truncation costs
more far-VRU recall than the energy it returns, so we forgo that energy by design. \emph{(iv)~Not every dense
task.} Detection, occupancy and segmentation all transfer, but monocular metric depth does not, because one
model's metric head changes sign under truncation; we therefore scope the claim to the three reported tasks. The
motion-compensated projection also needs per-dataset ego-pose and falls back to a static RoI where pose is
unavailable.

A per-detection importance field, built from data-driven class sensitivity with a VRU safety floor, routes each
$16{\times}16$ block to a semantic tier and precision group (near-lossless $k_{\mathrm{obj}}{=}7$
object tiers, $k_{\mathrm{bg}}{=}5$ context/background, $k_{\mathrm{sky}}{=}2$ sky), each coded by a selective-bit-dropping
primitive with Gray coding and a transition-aware flip; detections at $t$ are projected with ego-compensated
constant velocity to form the prior at $t{+}1$. Over $29$ detectors and $5$ occupancy and $5$ segmentation models
on $12$ datasets across five continents, \mbox{MotiMem-$\Omega$} lowers front-camera bit-1 density by about
$52\%$ and switching by about $48\%$, which the analytical model maps to a $36\%$ memory-interface energy
reduction, with a lower end of $27\%$ across the swept literature coefficients. It improves the joint
accuracy--energy operating point over the conference base and the energy-matched naive truncations, while image
codecs preserve accuracy only without reducing the memory-interface proxy. The advantage is especially visible in
dense per-pixel and per-voxel perception, consistent with the view that much background and sky fidelity is not
needed by the evaluated perception stack. An on-vehicle closed loop is a natural next step.

\FloatBarrier
\section*{Acknowledgment}
The authors received no specific grant funding for this work.

\section*{Conflict of Interest}
The authors declare that they have no conflicts of interest relevant to this work.

\end{document}